\documentclass[10pt,twocolumn,letterpaper]{article}

\usepackage[pagenumbers]{cvpr}     
\usepackage{booktabs} 
\usepackage{multirow}
\usepackage{colortbl} 
\newcommand{\sd}[1]{\textsubscript{\scriptsize#1}}   
\newcommand{\best}[1]{\textbf{#1}}             
\newcommand{\second}[1]{\underline{#1}}        
\definecolor{oursrow}{gray}{0.92}

\usepackage{multirow}

\definecolor{cvprblue}{rgb}{0.21,0.49,0.74}
\usepackage[breaklinks,colorlinks,allcolors=cvprblue]{hyperref}

\title{Beyond Sparse Prompts: Reference-Guided Progressive Prototype-Prompted SAM for Vessel Segmentation}

\author{
Donghang Lyu$^{*}$ \quad
Zichen Zhang$^{*}$ \quad
Oleh Dzyubachyk \quad
Marius Staring\\
Leiden University Medical Center
}

\begin{document}
\maketitle
\footnotetext{* Equal contribution.}
\begin{abstract}

Vessel segmentation across diverse anatomies and imaging modalities remains challenging due to complex vascular morphologies, large-scale variations, and low contrast. Although foundation models such as the Segment Anything Model (SAM) have shown promising generalization, their conventional sparse prompting at the initial decoding stage provides limited guidance for recovering fine vascular structures. Moreover, the original decoder struggles to preserve tiny vessels and fine-grained details. In this work, we go beyond conventional sparse prompting by proposing reference-guided progressive prototype-prompted SAM (RPP-SAM) for 2D vessel segmentation. It incorporates vascular prompt embeddings into every upsampling stage through a dedicated attention mechanism, enabling progressive refinement of multi-scale vascular features during decoding. However, vascular prototypes typically require ground-truth masks, which are unavailable during inference. To address this, we propose reference-guided vascular prototype learning, which estimates query-specific prototypes from top-$k$ vascular reference graphs retrieved from a database, eliminating the need for query mask annotations at inference. Together, these components enable fully automatic segmentation while preserving fine-grained vascular characteristics. Extensive experiments on 20 public datasets across multiple anatomies and imaging modalities demonstrate that RPP-SAM consistently outperforms existing general medical and vessel-specific segmentation foundation models, improving the overall Dice and IoU by 2.78 and 4.45 percentage points on average, respectively. Code and model weights will be publicly released.

\end{abstract}    
\section{Introduction}
\label{sec:intro}

Vessel segmentation plays an important role in clinical diagnosis and treatment planning, enabling the extraction of vascular structures from medical images~\cite{kock2024suitability,brugnara2023deep}. Compared with organ segmentation, it is more challenging because of the thin, elongated, and highly branching vascular morphology, substantial diameter variations, low contrast, and imaging artifacts. Considerable variations in vascular appearance across anatomies and imaging modalities further compound these challenges. Traditional methods are essentially image filters~\cite{frangi1998multiscale} that rely on prior knowledge of vascular morphology and often struggle to generalize beyond specific scenarios. In recent years, deep learning methods have substantially advanced vessel segmentation by learning complex vascular patterns, with applications spanning fundus imaging~\cite{wu2019vessel,wang2020hard,ma2024improved,luo2025pa}, X-ray angiography~\cite{iyer2021angionet,zhu2021coronary,he2025deep}, and optical coherence tomography angiography~\cite{liu2022disentangled,li2025direction}. Nevertheless, most existing methods are still designed for individual datasets or modalities. Although foundation models for medical image segmentation have attracted increasing attention, they are not specifically tailored to vessels. Only few foundation model-based approaches~\cite{song2025optimized} address 2D vessel segmentation across anatomies and modalities, leaving this task largely underexplored. Moreover, existing approaches lack effective mechanisms to explicitly capture discriminative modality-specific vascular characteristics, such as distinct branching patterns, and preserve fine details of tiny vessels.

\begin{figure}[!t]
\centering
\includegraphics[width=\columnwidth]{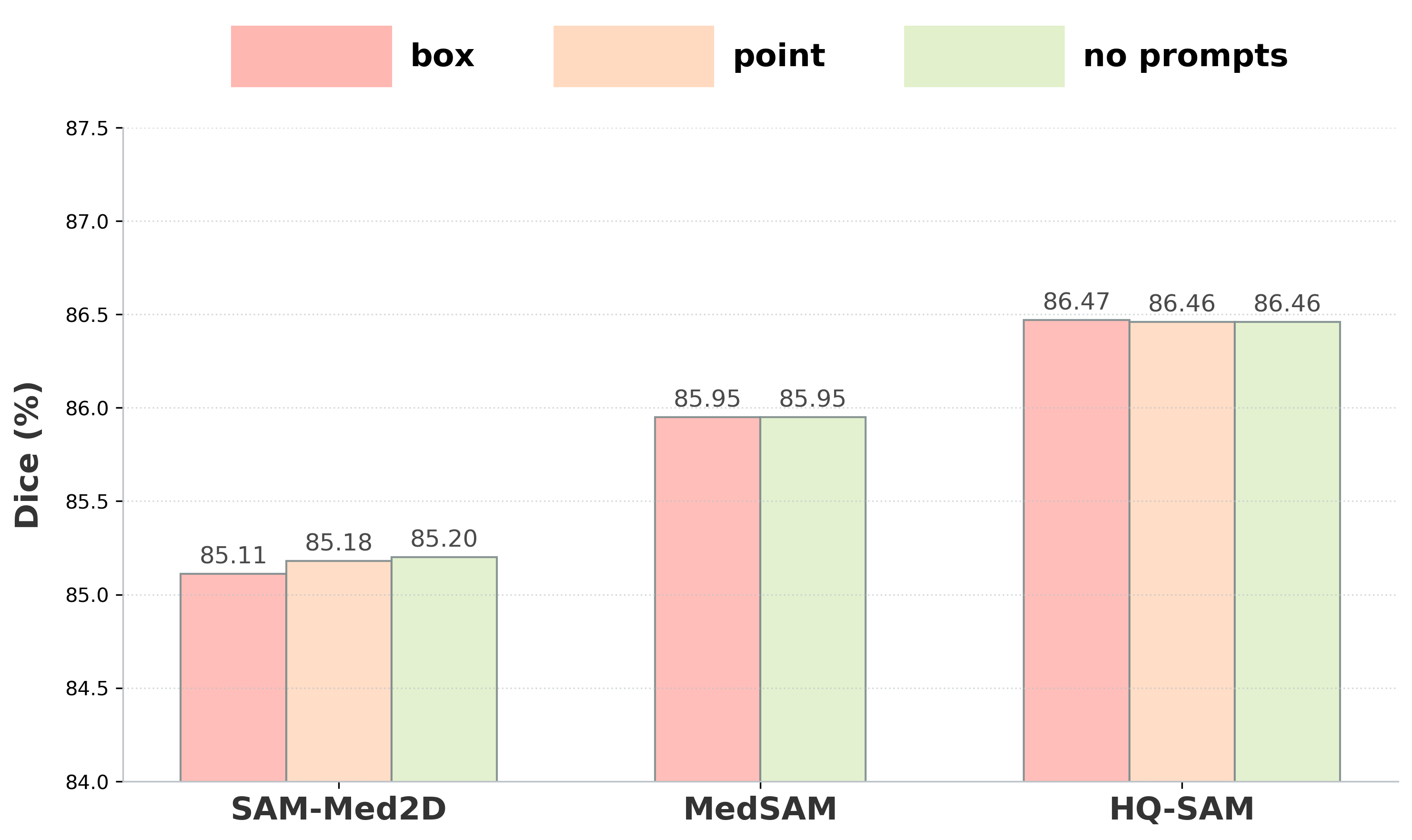}
\caption{A pilot study evaluating sparse prompting in representative SAM-based vessel segmentation methods shows that it has a negligible contribution to segmentation performance.}
\label{fig:intro_study}
\end{figure}

The Segment Anything Model (SAM)~\cite{kirillov2023segment}, a foundation model originally developed for natural image segmentation, has been successfully adapted to medical imaging~\cite{cheng2023sam,ma2024segment}, making it a promising backbone for general vessel segmentation. In SAM, the prompt encoder provides spatial guidance to the mask decoder through prompts such as points and bounding boxes. However, these sparse prompts are poorly suited to vascular structures. Unlike compact organs, vessels are elongated, highly branched, tree-like structures, making point prompts both difficult to place reliably and labor-intensive to annotate. Bounding boxes are similarly ineffective, as the widespread distribution of vessels often requires boxes to encompass nearly the entire field of view, providing limited spatial guidance. To further investigate their effectiveness, we conducted a pilot study comparing representative SAM-based methods with and without sparse prompts under same training conditions, with performance evaluated on our test set. The results show that sparse prompting offers negligible benefits for vessel segmentation, as shown in~\cref{fig:intro_study}. Beyond prompt limitations, the SAM's mask decoder, originally designed for generic object segmentation, struggles to preserve fine vascular details, such as thin branches and vessel bifurcations. This is further exacerbated by substantial inter-modality appearance variations, calling for modality-aware guidance during decoding. These observations highlight three key challenges in adapting SAM to vessel segmentation: \emph{designing informative vascular prompts, integrating them effectively into progressive decoding, and enhancing the decoder for multi-scale feature refinement to preserve fine-grained, modality-specific vascular structures.}

In this paper, we present reference-guided progressive prototype-prompted SAM (\textbf{RPP-SAM}), a fully automatic framework for 2D vessel segmentation across multiple imaging modalities and anatomies, designed to address the aforementioned challenges. Our contributions are summarized as follows:

\textbf{1. Progressive prototype-prompted decoding}. We introduce a new prompting paradigm that moves beyond conventional sparse prompts by injecting prototype-informed vascular prompt embeddings (VPEs) into progressive upsampling stages, providing fine-grained guidance throughout decoding. Specifically, we develop a vascular prototype attention (VPA) mechanism that enables effective interactions between VPEs and multi-scale decoder features, enhancing the recovery of fine-grained vascular structures.

\textbf{2. Reference-guided vascular prototype learning.} To obtain vascular prompts, we use a convolutional neural network (CNN) to extract multi-scale features from the query image and combine them with the interpolated ground-truth mask to generate a mask-derived vascular prototype, providing modality-specific vascular semantic information. However, these prototypes require ground-truth vascular masks, which are unavailable at inference. Therefore, we propose reference-guided vascular prototype learning. Specifically, we retrieve the top-$k$ vascular reference graphs from a database based on image-level similarity and encode their structural information into a reference-derived vascular prototype. During training, dedicated alignment losses align the two prototypes, which are then dynamically fused to form the VPE, progressively bridging the training-inference gap. At inference, only the reference-derived vascular prototype is used, eliminating the need for query ground-truth masks.

\textbf{3. Comprehensive cross-anatomy and cross-modality evaluation.} We establish a comprehensive benchmark comprising 20 public datasets across multiple imaging modalities and anatomies. Extensive experiments demonstrate that RPP-SAM consistently outperforms existing SAM-based foundation models for 2D vessel segmentation, including those equipped with manual prompts. Notably, these gains are particularly pronounced for fine-grained vascular structures, highlighting the effectiveness of our prompt design and progressive decoding strategy in recovering them.

\section{Related work}
\label{sec:related_work}

\subsection{Domain-specific vessel segmentation}

Early vessel segmentation methods~\cite{frangi1998multiscale,staal2004ridge} rely on handcrafted vascular priors, such as matched filtering, ridge detection, and Hessian-based vessel filtering, to enhance curvilinear structures. However, they require careful parameter tuning and often generalize poorly across anatomies and imaging modalities. Although deep learning has substantially advanced vessel segmentation, most existing methods~\cite{wu2019vessel,wang2020hard,ma2024improved,luo2025pa,iyer2021angionet,zhu2021coronary,he2025deep,liu2022disentangled,li2025direction} remain domain-specific and typically target individual datasets or imaging modalities. To alleviate the sparsity of vascular annotations, some approaches explore data synthesis~\cite{ma2021self,kim2022diffusion,lin2023yolocurvseg,zhang2024self,kreitner2024synthetic,zhang2024xa,liang2026self,hao2026dosta} to augment training data and improve segmentation performance. Beyond image-based learning, graph representations have also been explored for vessel segmentation. Existing approaches construct graphs from intermediate segmentation masks~\cite{shin2019deep,mishra2021vtg}, build graph structures within feature representations~\cite{xu2024g2vit,ahmed2026tffm}, or derive graphs from input images through patch-based partitioning~\cite{jalali2024vga,li2026learning}.

\subsection{Medical segmentation foundation models}

Medical segmentation foundation methods have largely built upon SAM framework by adapting it to medical domains through fine-tuning. SAM-Med2D~\cite{cheng2023sam} and MedSAM \cite{ma2024segment} are pioneering efforts in this direction. Subsequent work has focused on improving some specific aspects of SAM, such as parameter-efficient adaptation using adapters~\cite{wu2025medical,tejero2025sam} and enhancing computational efficiency~\cite{ma2024efficient,lyu2024mcp}. Despite these advances, foundation model-based 2D vessel segmentation remains largely underexplored. OVS-Net~\cite{song2025optimized} represents a recent state-of-the-art approach to general vessel segmentation, employing a hybrid encoder that combines a SAM backbone with a convolutional branch. However, it remains a semi-automatic framework that relies on conventional sparse point prompts and does not explicitly model modality-specific vascular representations.

\section{Methodology}

\begin{figure*}[!t]
\centering
\includegraphics[width=\linewidth]{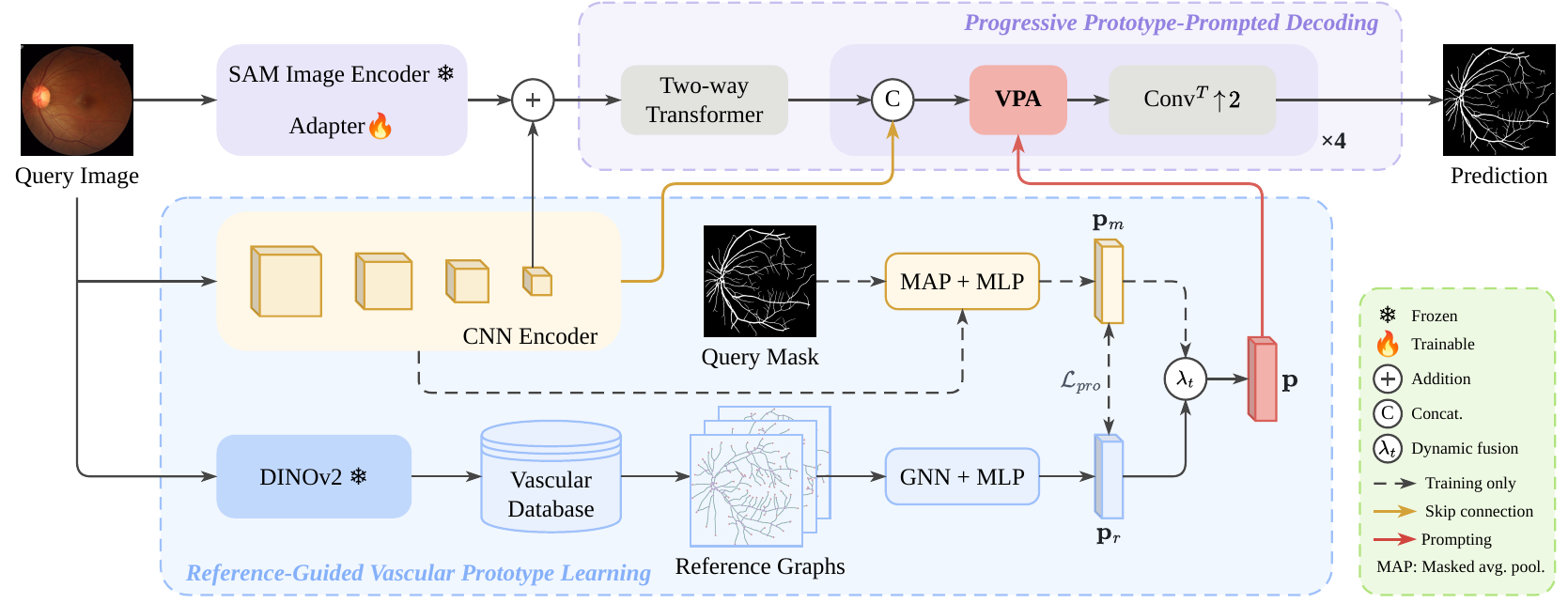}
\caption{Overview of the proposed RPP-SAM framework. Retrieved vascular graphs are encoded into a reference-derived vascular prototype, while a parallel CNN branch derives a mask-derived prototype from multi-scale features and the ground-truth mask. During training, the two prototypes are dynamically fused to form the vascular prompt embedding (VPE), which guides progressive decoding. At inference, only the reference-derived prototype is used as the VPE.}
\label{fig:framework}
\end{figure*}

In this section, we present the architecture and learning procedure of RPP-SAM for 2D vessel segmentation, as illustrated in~\cref{fig:framework}. The framework consists of three interconnected components: vascular reference retrieval and encoding, reference-guided vascular prototype learning, and progressive prototype-prompted decoding. Specifically, retrieved reference graphs provide structural cues for constructing a reference-derived vascular prototype, while multi-scale CNN features and ground-truth masks generate the mask-derived vascular prototype during training. We align these two representations and dynamically fuse them to form the VPE, which guides feature refinement throughout the progressive decoding process. At inference, the VPE is derived solely from the reference-derived prototype, eliminating the need for query ground-truth masks and enabling fully automatic segmentation.

\subsection{Reference vascular graph retrieval}

In general, vascular images from the same modality often share spatial and semantic characteristics, making modality-aware retrieval beneficial for vessel segmentation. Meanwhile, graph representations explicitly encode vascular geometry and connectivity, providing a structured representation of vessel morphology. We therefore construct a modality-organized vascular database containing images, their corresponding vascular graphs, and modality labels. Each vascular graph $G_{i}=(V_{i}, E_{i})$ is derived from the corresponding segmentation mask $\mathbf{M}_{i} \in \{ 0,1 \}^{h\times w}$, where $V_{i}$ and $E_{i}$ denote the node and edge sets, respectively, and $h$ and $w$ are spatial dimensions. Specifically, the mask is first skeletonized~\cite{zhang1984fast,soille1999morphological,lee1994building} and pruned to remove short, noisy branches. Skeleton pixels are then treated as nodes, local neighborhood connections as edges, and pixel coordinates as node features, preserving vessel geometry.

\begin{figure}[!t]
\centering
\includegraphics[width=\columnwidth]{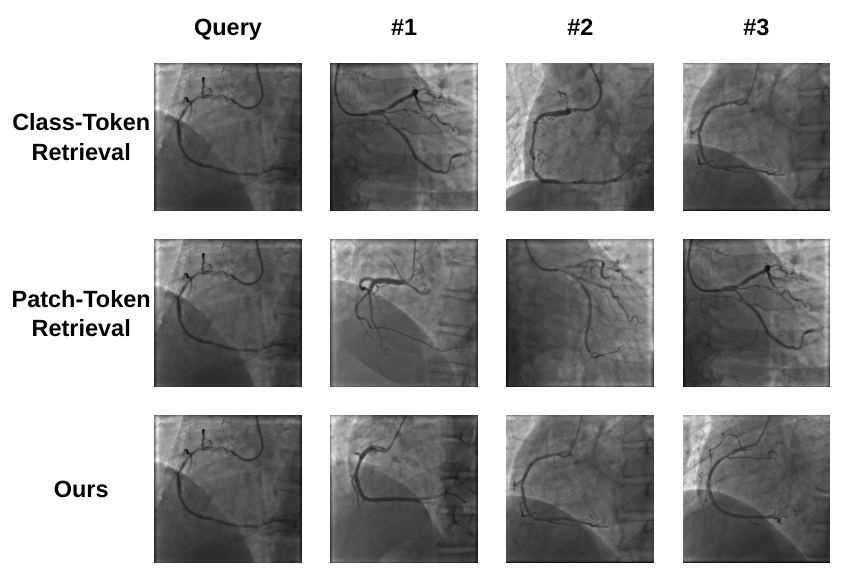}
\caption{A qualitative comparison of different retrieval methods on a coronary angiogram. The second to fourth columns show the top 3 retrieved images in descending order of similarity.}
\label{fig:retrieval}
\end{figure}

Given a query image $\mathbf{I}_{q} \in \mathbb{R}^{3 \times h \times w}$ with modality $m_{q}$, we first select images sharing the same modality from the database. We then use DINOv2~\cite{oquab2023dinov2,darcet2024vision} to extract patch tokens $\mathbf{T}_{q},\mathbf{T}_{i}\in\mathbb{R}^{h_{t}\times w_{t} \times c_{t}}$ for $\mathbf{I}_{q}$ and each candidate image $\mathbf{I}_{i}$. Instead of using the global $\texttt{[CLS]}$ token, we compute the image similarity $s$ as the mean cosine similarity between corresponding patch tokens:
\begin{equation}
s(\mathbf{I}_{q},\mathbf{I}_{i}) = \frac{1}{h_{t} w_{t}}
\sum_{u=1}^{h_{t}}\sum_{v=1}^{w_{t}}
\frac{\mathbf{T}_{q}(u,v)^\top \mathbf{T}_{i}(u,v)
}{
\|\mathbf{T}_{q}(u,v)\|_2
\|\mathbf{T}_{i}(u,v)\|_2
}.
\label{eq:image_similarity}
\end{equation}
This better accounts for spatial alignment, enabling us to select the top-$k$ most similar images, indexed by $N_{q}$, and then retrieve their corresponding vascular graphs $\{G_{i}\mid i\in N_{q}\}$.  \cref{fig:retrieval} qualitatively compares retrieval results using different strategies based on the \texttt{[CLS]} token and spatial patch tokens. Notably, the second row first averages all patch tokens before computing the similarity score.

\subsection{Progressive prototype-prompted decoding}

Our pilot study (\cref{fig:intro_study}) has revealed that sparse prompts have negligible contributions when used as initial inputs to the mask decoder for vessel segmentation. This motivates rethinking both the prompting strategy and decoder architecture to better exploit prompt-based guidance for vessel segmentation. We therefore introduce informative vascular prototype prompts throughout decoding and adapt the decoder architecture to integrate them. Specifically, the original mask decoder lacks skip connections from the image encoder and employs only two upsampling stages, followed by bilinear interpolation to recover the input resolution. These limitations hinder the preservation of fine-grained vascular structures, which require rich local features. Therefore, we incorporate skip connections from a parallel CNN encoder to provide multi-scale local features and introduce four progressive upsampling stages to produce original-resolution predictions directly. In addition, the final feature maps from the CNN branch are added to the SAM image embeddings, providing complementary spatial guidance analogous to dense prompting.

\begin{figure}[!t]
\centering
\includegraphics[width=\columnwidth]{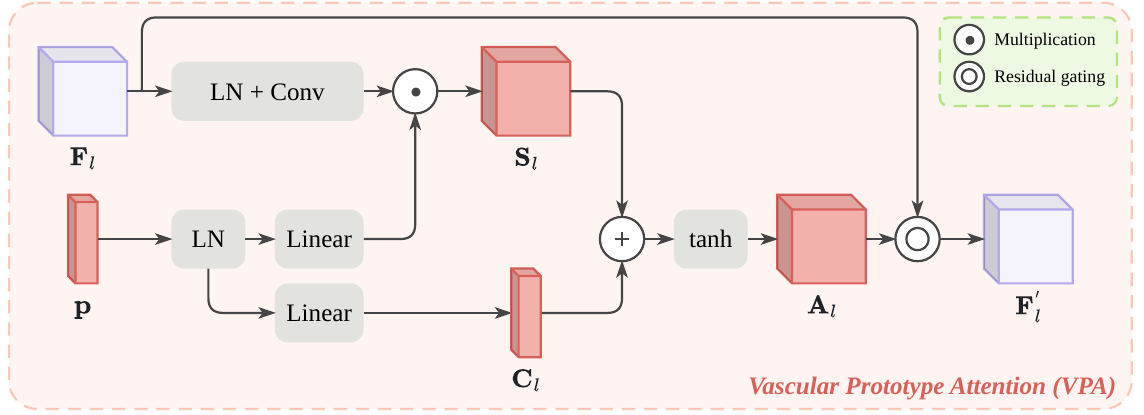}
\caption{Overview of vascular prototype attention (VPA).}
\label{fig:vpa}
\end{figure}

Beyond architectural refinement, we introduce vascular prototype attention (VPA), as shown in~\cref{fig:vpa}, to progressively inject vascular prompt embeddings (VPEs) into each upsampling stage for modality-aware guidance throughout decoding. Given a VPE $\mathbf{p}\in\mathbb{R}^{c_{p}}$ and the feature map $\mathbf{F}_{l}\in\mathbb{R}^{c_{l}\times h_{l}\times w_{l}}$ at stage $l$, we first project them to align the channel dimension:
\begin{equation}
  \hat{\mathbf{F}}_{l} = \mathrm{Conv}(\mathrm{LN}(\mathbf{F}_{l})),\qquad \hat{\mathbf{p}}_{l} = \mathrm{Linear}(\mathrm{LN}(\mathbf{p})),
  \label{eq:f_l}
\end{equation}
where $\hat{\mathbf{F}}_{l} \in \mathbb{R}^{c_{l} \times h_{l} \times w_{l}}$ and $\hat{\mathbf{p}}_{l} \in \mathbb{R}^{c_{l}}$ denote the projected $\mathbf{F}_{l}$ and $\mathbf{p}$, respectively, $\mathrm{LN}$ is layer normalization. A scaled dot product is then computed between $\hat{\mathbf{p}}_{l}$ and $\hat{\mathbf{F}}_{l}$, yielding a spatial score map $\mathbf{S}_{l} \in \mathbb{R}^{1 \times h_{l} \times w_{l}}$:
\begin{equation}
  \mathbf{S}_{l} = \frac{\hat{\mathbf{p}}_{l}^{\top} \hat{\mathbf{F}}_{l}}{\sqrt{c_{l}}}.
  \label{eq:s_l}
\end{equation}
For channel-wise modulation, we apply another linear projection to $\mathbf{p}$ to obtain channel-wise scores $\mathbf{C}_{l} \in \mathbb{R}^{c_{l} \times 1 \times 1}$:
\begin{equation}
  \mathbf{C}_{l} = \mathrm{Linear}(\mathrm{LN}(\mathbf{p})).
  \label{eq:c_l}
\end{equation}
The spatial and channel-wise scores are then combined through broadcasting and passed through a $\tanh$ activation to obtain the joint attention map $\mathbf{A}_{l} \in \mathbb{R}^{c_{l} \times h_{l} \times w_{l}}$:
\begin{equation}
  \mathbf{A}_{l} = \tanh(\mathbf{S}_{l} + \mathbf{C}_{l}),
  \label{eq:a_l}
\end{equation}
Finally, we apply the attention map to the original feature map $\mathbf{F}_{l}$ through residual gating:
\begin{equation}
  \mathbf{F}_{l}^{'} = \mathbf{F}_{l} \odot \left(1 + \alpha \mathbf{A}_{l}\right),
  \label{eq:f_l0}
\end{equation}
where $\alpha$ controls the strength of attention modulation and $\odot$ denotes element-wise multiplication.

\subsection{Reference-guided vascular prototype learning}

As described above, VPEs progressively guide the decoding process. We next detail how they are learned from query images and retrieved vascular graphs.

\textbf{Mask-derived vascular prototypes.} During training, we first extract multi-scale feature maps from the query image using the aforementioned ResNet-style~\cite{he2016deep} CNN encoder. Unlike vision transformers~\cite{dosovitskiy2020image}, which typically operate on image tokens at a uniform resolution, CNNs naturally learn hierarchical representations with rich local information, making them well suited for learning vascular prototypes. Given a feature map  $\mathbf{F}_{l} \in \mathbb{R}^{c_{l}\times h_{l}\times w_{l}}$ at stage $l$, we resize the corresponding ground-truth mask to the same spatial resolution through nearest-neighbor interpolation, yielding $\mathbf{M}_{l} \in \{ 0,1 \}^{h_{l} \times w_{l}}$. We then obtain a vascular representation $\mathbf{p}_{l} \in \mathbb{R}^{c_{l}}$ for this scale through masked average pooling:
\begin{equation}
\mathbf{p}_{l} = \frac{\sum_{x,y} \mathbf{F}_{l}(x,y) \cdot \mathbf{M}_{l}(x,y)}{\sum_{x,y} \mathbf{M}_{l}(x,y)}.
\label{eq:p_l}
\end{equation}
The resulting multi-scale representations $\left \{ \mathbf{p}_{l} \right \}_l$ are concatenated and fed into a multi-layer perceptron (MLP) to produce the mask-derived vascular prototype $\mathbf{p}_{\mathrm{m}} \in \mathbb{R}^{c_{p}}$, which encodes vascular appearance characteristics.

\textbf{Reference-derived vascular prototypes.} To estimate query-specific vascular prototypes without relying on ground-truth masks at inference, we leverage the retrieved vascular graphs to derive reference-derived prototype embeddings and align them with their mask-derived counterparts $\mathbf{p}_{\mathrm{m}}$ during training. Specifically, we employ a transformer-based graph neural network (GNN), denoted by $\mathrm{G}$, to encode each retrieved graph. It consists of an input projection layer followed by multiple TransformerConv~\cite{shi2020masked} blocks to capture vascular structural patterns through graph message passing. The resulting node embeddings are refined by linear layers and aggregated via global average pooling (GAP) to obtain a graph-level embedding $\mathbf{g}_{i} \in \mathbb{R}^{c_{g}}$ for each graph $G_{i}$:
\begin{equation}
  \mathbf{g}_{i} = \mathrm{GAP}(\mathrm{G}(G_{i})), i \in N_{q}.
  \label{eq:g_i}
\end{equation}
Given the top-$k$ graph-level embeddings, we concatenate them and feed the resulting representation into an MLP to obtain a joint embedding $\mathbf{p}_{\mathrm{r}} \in \mathbb{R}^{c_{p}}$:
\begin{equation}
  \mathbf{p}_{\mathrm{r}} = \mathrm{MLP} \left(\mathrm{concat}(\{\mathbf{g}_{i} \mid i \in N_{q}\})\right).
  \label{eq:p_g_i}
\end{equation}
By aggregating structural information from multiple retrieved graphs, $\mathbf{p}_{\mathrm{r}}$ encodes complementary vascular morphology and connectivity patterns, providing structural priors for learning query-specific vascular prototypes.

\textbf{Prototype alignment and inference.} During training, we align the reference-derived prototype $\mathbf{p}_{\mathrm{r}}$ with its corresponding mask-derived prototype $\mathbf{p}_{\mathrm{m}}$ through alignment losses. To progressively shift the decoding guidance from mask-derived to reference-derived representations, we employ a dynamic fusion strategy to construct the VPE:
\begin{equation}
\mathbf{p}
= (1 - \lambda_{t}) \mathbf{p}_{\mathrm{m}} + \lambda_{t}\mathbf{p}_{\mathrm{r}},
\label{eq:p}
\end{equation}
where $\lambda_{t}$ gradually increases from $0.5$ to $1$ during training. At inference, only $\mathbf{p}_{\mathrm{r}}$ is available, and thus directly serves as the VPE, equivalent to $\lambda_{t}=1$. This formulation ensures consistent prototype guidance across training and inference while enabling fully automatic vessel segmentation.

\subsection{Loss function}

The overall loss function consists of a segmentation loss and a prototype alignment loss. The segmentation loss $\mathcal{L}_{\mathrm{seg}}$ is defined as:
\begin{equation}
  \mathcal{L}_{\mathrm{seg}} = \mathcal{L}_{\mathrm{Dice}} + \mathcal{L}_{\mathrm{BCE}} + \lambda_{\mathrm{clDice}} \mathcal{L}_{\mathrm{clDice}},
  \label{eq:l_seg}
\end{equation}
where $\mathcal{L}_{\mathrm{clDice}}$ is the clDice loss~\cite{shit2021cldice}. To align $\mathbf{p}_{\mathrm{r}}$ with its mask-derived counterpart $\mathbf{p}_{\mathrm{m}}$, we employ an $\mathcal{L}_{1}$ loss and a contrastive loss~\cite{radford2021learning}. The $\mathcal{L}_{1}$ loss directly minimizes their embedding discrepancy, while the contrastive loss promotes contrastive alignment in the embedding space. The prototype alignment loss is formulated as:
\begin{equation}
  \mathcal{L}_\mathrm{pro} = \lambda_{1} \mathcal{L}_{1}(\mathbf{p}_{\mathrm{m}}, \mathbf{p}_{\mathrm{r}}) + \lambda_{2} \mathcal{L}_{\mathrm{con}}(\mathbf{p}_{\mathrm{m}},\mathbf{p}_{\mathrm{r}}).
  \label{eq:l_p}
\end{equation}
The overall training objective is then given by:
\begin{equation}
  \mathcal{L} = \mathcal{L}_{\mathrm{seg}} + \mathcal{L}_{\mathrm{pro}}.
  \label{eq:l}
\end{equation}

\section{Experiments}

\subsection{Datasets}

We collected 20 public datasets spanning four imaging modalities to establish a comprehensive benchmark, with details summarized in~\cref{tab:dataset}. After excluding Fundus-AVSeg~\cite{deng2025fundus} for zero-shot unseen dataset generalization, we merged the remaining datasets and split them into training and test sets at an 80:20 ratio, yielding 3121 and 785 samples, respectively. For datasets containing multiple samples per patient, we performed patient-disjoint splitting to prevent data leakage. Fundus-AVSeg~\cite{deng2025fundus}, comprising 100 samples, was held out as a zero-shot test set. All images were converted to three channels, resized to $1024 \times 1024$ using bicubic interpolation, and normalized to $\left [ 0,1 \right ]$. During training, we applied random contrast and brightness adjustments, gamma correction, and Gaussian noise for data augmentation.

\begin{table}[t]
\caption{A summary of the datasets. OCTA and DSA are optical coherence tomography angiography and digital subtraction angiography, respectively.}
\label{tab:dataset}
\centering
\small
\setlength{\tabcolsep}{3.6pt}
\begin{tabular}{ccc}
\toprule
Dataset & Modality & Sample Number \\
\midrule
ARIA~\cite{farnell2008enhancement} & Fundus & 143 \\
CHASE-AV~\cite{fraz2012ensemble} & Fundus & 28 \\
DRIVE~\cite{staal2004ridge} & Fundus & 40 \\
FIVES~\cite{jin2022fives} & Fundus & 797 \\
FOVEA~\cite{ravasio2025fovea} & Fundus & 40 \\
HRF~\cite{budai2013robust} & Fundus & 45 \\
LES-AV~\cite{orlando2018towards} & Fundus & 22 \\
STARE~\cite{hoover2000locating} & Fundus & 20 \\
OCTA-500 3mm~\cite{li2024octa} & OCTA & 200 \\
OCTA-500 6mm~\cite{li2024octa} & OCTA & 300 \\
ROSE-1~\cite{ma2020rose} & OCTA & 39 \\
ROSSA~\cite{ning2024accurate} & OCTA & 918 \\
DIAS~\cite{liu2024dias} & DSA & 60 \\
DSCA~\cite{zhang2025dsca} & DSA & 224 \\
DCA1~\cite{cervantes2019automatic} & X-Ray (Coronary) & 134 \\
FS-CAD~\cite{zeng2024pretrained} & X-Ray (Coronary) & 50 \\
XACV~\cite{wu2025denver} & X-Ray (Coronary) & 221 \\
XCAD~\cite{ma2021self} & X-Ray (Coronary) & 126 \\
Dr-SAM~\cite{zohranyan2024dr} & X-Ray (Peripheral) & 499 \\
\midrule
Fundus-AVSeg~\cite{deng2025fundus} & Fundus & 100 \\
\bottomrule
\end{tabular}
\end{table}

\subsection{Implementation details}

Given the limited training data, we adopted adapters~\cite{chen2022adaptformer} in each transformer block of the SAM image encoder for parameter-efficient adaptation to vascular images. Only the adapters and the final transformer block were optimized, while the remaining image encoder parameters were frozen. All the other components were trainable. We initialized the image encoder with pretrained MedSAM-B weights~\cite{ma2024efficient}.

RPP-SAM was implemented in PyTorch 2.6.0 and trained on a single NVIDIA A100 GPU (80 GB memory) for 25 epochs using AdamW with a learning rate of $1\times10^{-4}$ and a batch size of 2. The loss coefficients $\lambda_{\mathrm{clDice}}, \lambda_{1}, \lambda_{2}$ were all set to 0.1. The number of retrieved reference graphs, $k$, was set to 3 based on empirical evaluation.

We compared RPP-SAM with six representative methods. SAM-Med2D~\cite{cheng2023sam} and MedSAM~\cite{ma2024segment} are widely used foundation models for medical image segmentation, while HQ-SAM~\cite{ke2023segment} is designed for high-fidelity segmentation and serves as a strong baseline. We further included ASPS~\cite{li2024asps}, a polyp segmentation method with potential applicability to vessel segmentation; OVS-Net~\cite{song2025optimized}, a recent advanced method featuring a dual-branch architecture for general 2D vessel segmentation; and SACM~\cite{zhu2026dual}, a prompt-free, SAM-based framework that employs dual-level adapters for cross-domain curvilinear structure segmentation. Regarding prompting strategies, SAM-Med2D and HQ-SAM support both bounding-box and point prompts, so we evaluated two variants of each. For MedSAM and OVS-Net, we only followed their original designs, using bounding-box and point prompts, respectively, while ASPS and SACM are prompt-free. For a fair comparison, all methods were retrained on our benchmark dataset and initialized imaage encoder with MedSAM-B weights. Segmentation performance was evaluated using Dice, intersection over union (IoU), and clDice.

\subsection{Quantitative results}

\begin{figure}[!t]
\centering
\includegraphics[width=\columnwidth]{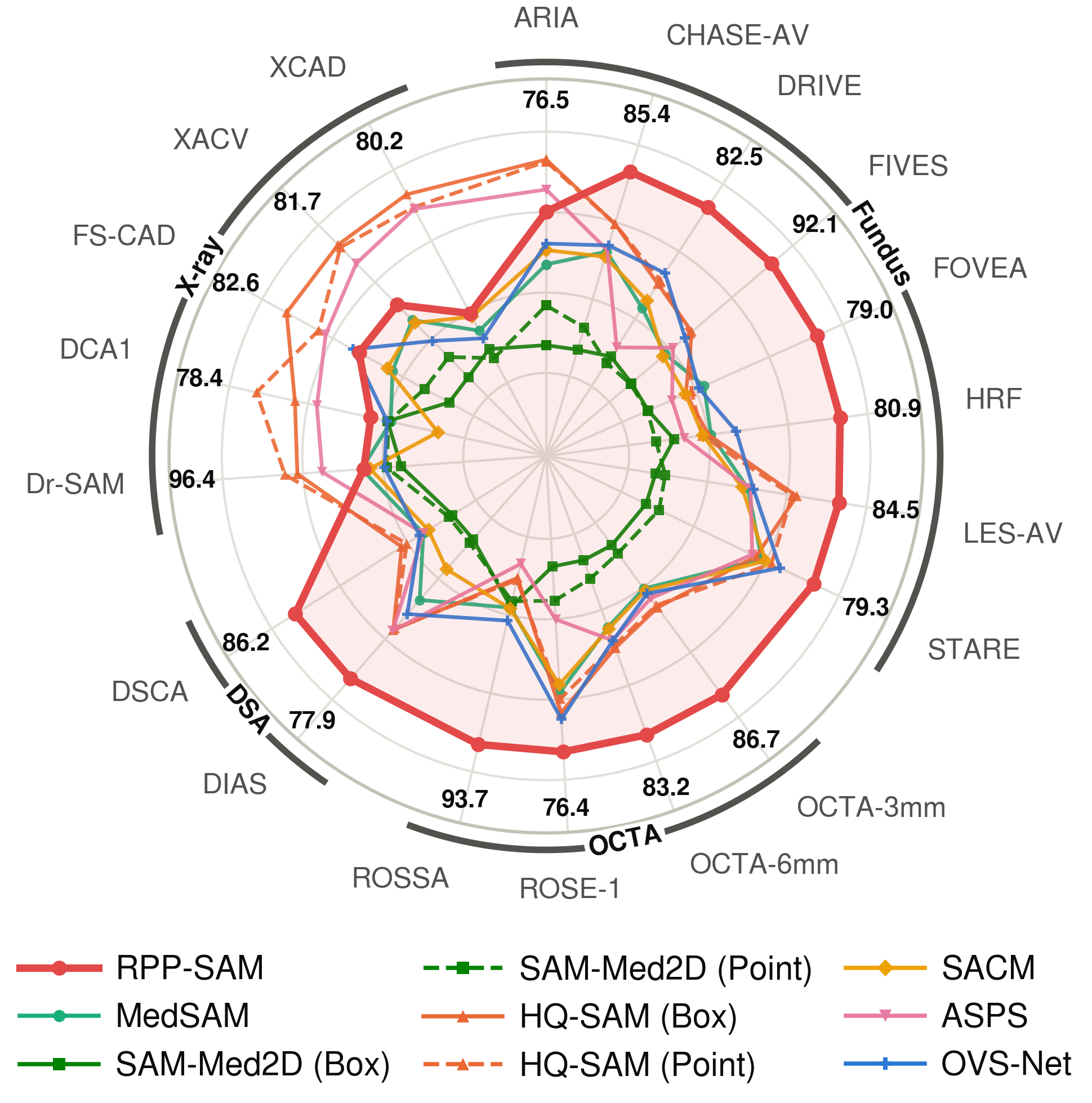}
\caption{Dataset-wise Dice scores across different methods.}
\label{fig:dataset_performance}
\end{figure}

Quantitative results in~\cref{tab:performance}, grouped by modality, demonstrate that RPP-SAM achieves the best overall segmentation performance across seen and zero-shot unseen datasets. Although it does not rank first in all modalities, it consistently excels in the three most challenging modalities: fundus, OCTA, and DSA, where abundant thin and highly branched vessels make accurate segmentation particularly challenging. In X-ray angiography, annotations primarily cover major vessels, leaving many small branches unlabeled. Meanwhile, background structures such as catheters and bone edges often exhibit tubular appearances similar to vessels. RPP-SAM may therefore respond more strongly to these unlabeled or vessel-like structures, producing additional false-positive fragments and consequently reducing its performance on this modality. Despite this limitation, RPP-SAM maintains a substantial performance margin over competing methods on the zero-shot test set, demonstrating strong cross-dataset generalization capability. These results are particularly encouraging given its fully automatic operation without manual prompting, highlighting its potential for clinical applications. The dataset-level comparison in~\cref{fig:dataset_performance} further demonstrates its robust and balanced performance, with leading results on the most challenging datasets and only modest performance drops on a few X-ray datasets, consistent with the modality-level findings.

\begin{table*}[t]
\caption{\textbf{Quantitative comparison on vessel segmentation.}
We report Dice, IoU, and clDice (\%, mean$_{\mathrm{std}}$; higher is better) for each imaging modality,
the average over all four modalities (\emph{Total}), and on unseen datasets (\emph{Zero-Shot}).
\emph{Prompt} denotes the prompt used at inference; \emph{Auto} requires no manual prompt.
\best{Bold} and \second{underlined} indicate the best and second-best results.}
\label{tab:performance}
\centering
\small
\setlength{\tabcolsep}{3.0pt}
\begin{tabular}{@{}ll ccc ccc ccc@{}}
\toprule
\multirow{2}{*}{Method} & \multirow{2}{*}{Prompt}
& \multicolumn{3}{c}{Fundus} & \multicolumn{3}{c}{OCTA} & \multicolumn{3}{c}{DSA} \\
\cmidrule(lr){3-5} \cmidrule(lr){6-8} \cmidrule(l){9-11}
& & Dice$\uparrow$ & IoU$\uparrow$ & clDice$\uparrow$ & Dice$\uparrow$ & IoU$\uparrow$ & clDice$\uparrow$ & Dice$\uparrow$ & IoU$\uparrow$ & clDice$\uparrow$ \\
\midrule
SAM-Med2D~\cite{cheng2023sam} & Box & 83.61\sd{6.95} & 72.42\sd{9.92} & 83.70\sd{8.64} & 86.08\sd{5.90} & 76.01\sd{8.58} & 90.50\sd{6.27} & 77.03\sd{5.12} & 62.92\sd{6.67} & 72.54\sd{6.02} \\
SAM-Med2D~\cite{cheng2023sam} & Point & 83.64\sd{6.87} & 72.46\sd{9.81} & 83.80\sd{8.61} & 86.12\sd{5.64} & 76.03\sd{8.21} & 90.90\sd{6.04} & 77.20\sd{5.13} & 63.14\sd{6.68} & 72.31\sd{6.19} \\
MedSAM~\cite{ma2024segment} & Box & 84.95\sd{6.66} & 74.39\sd{9.68} & 85.63\sd{8.16} & 86.67\sd{5.20} & 76.82\sd{7.61} & 91.81\sd{5.76} & 78.63\sd{4.91} & 65.05\sd{6.55} & 74.54\sd{6.06} \\
HQ-SAM~\cite{ke2023segment} & Box & \second{85.88}\sd{6.54} & \second{75.80}\sd{9.63} & \second{86.67}\sd{8.26} & 86.31\sd{4.68} & 76.20\sd{6.83} & \best{91.98}\sd{5.53} & \second{79.72}\sd{4.54} & \second{66.51}\sd{6.17} & 74.83\sd{6.35} \\
HQ-SAM~\cite{ke2023segment} & Point & 85.82\sd{6.51} & 75.71\sd{9.59} & 86.40\sd{8.35} & 86.33\sd{4.74} & 76.23\sd{6.89} & \second{91.96}\sd{5.64} & 79.51\sd{4.76} & 66.24\sd{6.43} & 74.39\sd{6.50} \\
ASPS~\cite{li2024asps} & Auto  & 85.20\sd{6.59} & 74.76\sd{9.61} & 85.73\sd{8.58} & 85.94\sd{4.74} & 75.62\sd{6.78} & 91.62\sd{5.95} & 78.84\sd{4.55} & 65.30\sd{6.15} & 72.78\sd{6.29} \\
OVS-Net~\cite{song2025optimized} & Point & 85.58\sd{6.81} & 75.39\sd{9.97} & 86.46\sd{7.77} & \second{86.99}\sd{5.18} & \second{77.33}\sd{7.67} & 91.93\sd{5.50} & 78.87\sd{4.82} & 65.37\sd{6.49} & \second{75.98}\sd{6.23} \\
SACM~\cite{zhu2026dual} & Auto  & 84.88\sd{6.69} & 74.29\sd{9.69} & 85.83\sd{8.16} & 86.69\sd{5.18} & 76.86\sd{7.59} & 91.87\sd{5.50} & 78.21\sd{5.11} & 64.51\sd{6.76} & 73.84\sd{6.58} \\
\rowcolor{oursrow}
RPP-SAM (Ours) & Auto & \best{88.35}\sd{7.40} & \best{79.88}\sd{11.35} & \best{89.07}\sd{7.53} & \best{90.10}\sd{5.87} & \best{82.47}\sd{9.15} & 90.85\sd{5.82} & \best{84.44}\sd{5.39} & \best{73.43}\sd{7.82} & \best{82.70}\sd{6.69} \\
\midrule[\heavyrulewidth]
\multirow{2}{*}{Method} & \multirow{2}{*}{Prompt}
& \multicolumn{3}{c}{X-Ray} & \multicolumn{3}{c}{Total} & \multicolumn{3}{c}{Zero-Shot} \\
\cmidrule(lr){3-5} \cmidrule(lr){6-8} \cmidrule(l){9-11}
& & Dice$\uparrow$ & IoU$\uparrow$ & clDice$\uparrow$ & Dice$\uparrow$ & IoU$\uparrow$ & clDice$\uparrow$ & Dice$\uparrow$ & IoU$\uparrow$ & clDice$\uparrow$ \\
\midrule
SAM-Med2D~\cite{cheng2023sam} & Box & 87.62\sd{10.20} & 79.30\sd{14.91} & 87.31\sd{11.05} & 85.11\sd{7.99} & 74.88\sd{11.65} & 86.37\sd{9.68} & 84.50\sd{2.44} & 73.23\sd{3.58} & 84.86\sd{3.57} \\
SAM-Med2D~\cite{cheng2023sam} & Point & 87.74\sd{10.17} & 79.49\sd{14.88} & 87.60\sd{10.90} & 85.18\sd{7.89} & 74.97\sd{11.51} & 86.61\sd{9.68} & 84.72\sd{2.50} & 73.57\sd{3.69} & 84.89\sd{3.60} \\
MedSAM~\cite{ma2024segment} & Box & 88.05\sd{10.00} & 79.95\sd{14.67} & 88.93\sd{10.63} & 85.95\sd{7.53} & 76.08\sd{11.07} & 87.99\sd{9.20} & 85.78\sd{2.44} & 75.18\sd{3.65} & 86.57\sd{3.64} \\
HQ-SAM~\cite{ke2023segment} & Box & \second{89.20}\sd{9.35} & \second{81.65}\sd{13.73} & \second{90.44}\sd{9.90} & \second{86.47}\sd{7.11} & \second{76.82}\sd{10.55} & \second{88.78}\sd{8.95} & 85.38\sd{2.55} & 74.57\sd{3.81} & 86.38\sd{3.81} \\
HQ-SAM~\cite{ke2023segment} & Point & \best{89.25}\sd{9.26} & \best{81.72}\sd{13.63} & \best{90.47}\sd{9.92} & 86.46\sd{7.12} & 76.80\sd{10.56} & 88.67\sd{9.08} & 85.60\sd{2.72} & 74.92\sd{4.07} & 86.23\sd{4.02} \\
ASPS~\cite{li2024asps} & Auto  & 88.93\sd{9.45} & 81.24\sd{13.88} & 89.95\sd{10.06} & 86.00\sd{7.23} & 76.10\sd{10.66} & 88.09\sd{9.40} & 84.87\sd{2.71} & 73.81\sd{3.99} & 85.45\sd{4.15} \\
OVS-Net~\cite{song2025optimized} & Point & 87.95\sd{10.04} & 79.79\sd{14.68} & 88.56\sd{10.76} & 86.24\sd{7.54} & 76.54\sd{11.10} & 88.28\sd{8.90} & \second{86.63}\sd{2.27} & \second{76.48}\sd{3.46} & \second{87.87}\sd{3.22} \\
SACM~\cite{zhu2026dual} & Auto & 87.94\sd{10.02} & 79.78\sd{14.72} & 88.52\sd{10.60} & 85.88\sd{7.57} & 75.98\sd{11.13} & 87.91\sd{9.22} & 85.87\sd{2.60} & 75.32\sd{3.89} & 86.36\sd{3.82} \\
\rowcolor{oursrow}
RPP-SAM (Ours) & Auto & 88.25\sd{9.88} & 80.23\sd{14.46} & 89.12\sd{10.64} & \best{88.69}\sd{7.66} & \best{80.47}\sd{11.55} & \best{89.28}\sd{8.14} & \best{89.60}\sd{2.03} & \best{81.23}\sd{3.26} & \best{91.99}\sd{2.88} \\
\bottomrule
\end{tabular}
\end{table*}

\subsection{Qualitative results}

\begin{figure*}[!t]
\centering
\includegraphics[width=\linewidth]{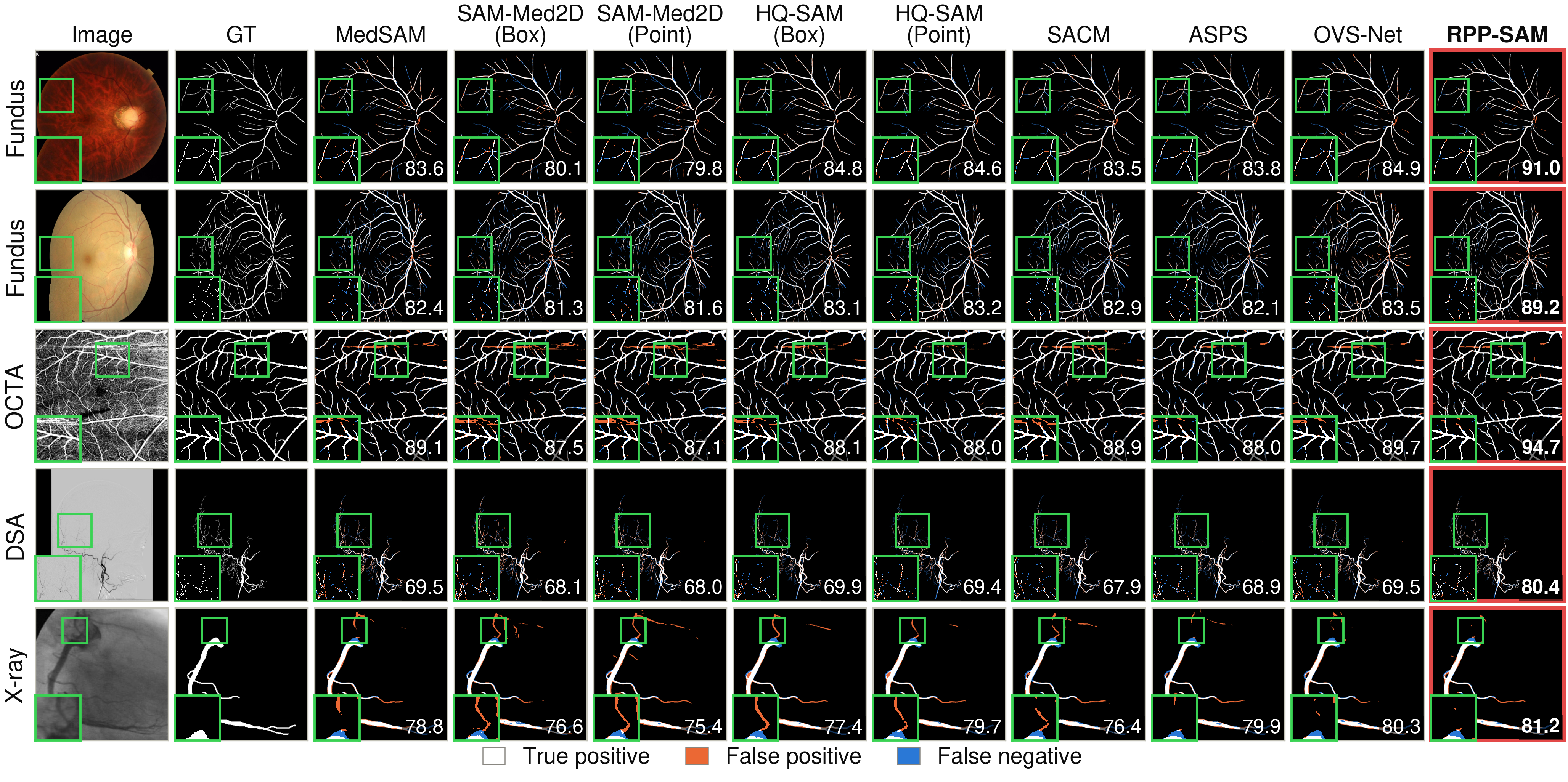}
\caption{Qualitative comparison across imaging modalities. From top to bottom: two fundus images (the second from the unseen Fundus-AVSeg test set), OCTA, DSA, and X-ray angiography. Columns show the input images, ground truths (GT), baseline predictions, and RPP-SAM predictions. Error maps indicate true positives (white), false positives (orange), and false negatives (blue). Green boxes indicate regions enlarged in the bottom-left insets to provide a closer view of fine details. Dice scores (\%) are reported in each panel, with the best score per row in bold.}
\label{fig:segmentation}
\end{figure*}

\cref{fig:segmentation} provides qualitative comparisons of RPP-SAM and baseline methods across different imaging modalities. As highlighted by the green boxes, baseline methods frequently fail to capture thin and low-contrast vessels, resulting in incomplete vascular structures and loss of fine morphological details. In contrast, RPP-SAM better preserves these delicate structures and captures complex vascular patterns, including fine branches and intricate vessel connections. Although some subtle vessel details remain challenging, RPP-SAM generally produces more structurally complete and morphologically faithful segmentation results, consistent with its strong generalization performance on the zero-shot unseen dataset. These observations highlight the effectiveness of our prompting strategy and decoding refinement in recovering fine-grained vascular structures across complex anatomies and diverse modalities.

\subsection{Ablation studies}

\textbf{Component designs.} We investigate the contributions of individual components in RPP-SAM through progressive ablation, with results summarized in~\cref{tab:ablation_studies1}. Starting from the original SAM architecture, we incrementally introduce the proposed components. \emph{CNN} adds a parallel CNN branch whose feature maps are fused with the SAM image embeddings. \emph{4 Upsampling} extends the decoder with two additional upsampling stages and incorporates CNN skip connections at all four stages for multi-scale feature fusion. \emph{VPE \& VPA} introduces VPEs through the proposed VPA mechanism at each upsampling stage. Finally, \emph{$\lambda_{t}$} enables dynamic prototype fusion to progressively align the training and inference settings.

Overall, performance consistently improves as components are added, demonstrating their complementary contributions and validating the overall architecture. These results further highlight the importance of progressive feature refinement and multi-scale fusion for recovering fine-grained vascular structures. Although dynamic fusion with $\lambda_{t}$ yields relatively modest gains, it facilitates a smooth transition between training and inference while promoting effective prototype alignment.

\begin{table}[t]
\caption{Ablation results for individual architectural components of RPP-SAM. Best results are highlighted in \best{bold}.}
\label{tab:ablation_studies1}
\centering
\small
\setlength{\tabcolsep}{3pt}
\begin{tabular}{@{}cccccccc@{}}
\toprule
\multicolumn{5}{c}{Architectures}
& \multicolumn{3}{c}{Metrics} \\
\cmidrule(lr){1-5} \cmidrule(lr){6-8}
SAM & CNN & 4 Up. &
\begin{tabular}{@{}c@{}}VPE\\\&VPA\end{tabular} &
$\lambda_{t}$ & Dice$\uparrow$ & IoU$\uparrow$ & clDice$\uparrow$ \\
\midrule
\checkmark &  &  &  &  & 86.40\sd{7.54} & 76.79\sd{11.15} & 88.55\sd{8.83} \\
\checkmark & \checkmark &  &  &  & 87.24\sd{7.66} & 78.14\sd{11.42} & \best{89.37}\sd{8.46} \\
\checkmark & \checkmark & \checkmark &  &  & 88.07\sd{7.97} & 79.53\sd{11.99} & 88.83\sd{8.22} \\
\checkmark & \checkmark & \checkmark & \checkmark &  & 88.65\sd{7.78} & 80.43\sd{11.71} & 89.23\sd{8.19} \\
\checkmark & \checkmark & \checkmark & \checkmark & \checkmark & \best{88.69}\sd{7.66} & \best{80.47}\sd{11.55} & 89.28\sd{8.14} \\
\bottomrule
\end{tabular}
\end{table}

\textbf{Alternative designs.} We further evaluate alternative configurations to validate our design choices, with results reported in~\cref{tab:ablation_studies2}. Replacing VPA with simple linear fusion, which directly adds VPEs to decoder feature maps, causes a performance drop. However, the degradation is relatively modest, as VPEs still provide useful semantic guidance during decoding. Nevertheless, it is less effective than VPA, which adaptively modulates spatial and channel-wise features. Furthermore, removing either prototype alignment loss weakens the alignment between mask-derived and reference-derived prototypes, degrading performance. This effect is more pronounced when removing $\mathcal{L}_{1}$, as the contrastive loss alone provides insufficient supervision under limited training data and small batch sizes. The resulting misalignment can lead to unreliable prompt embeddings that provide misleading guidance in the decoding process.

\begin{table}[t]
\caption{Ablation results comparing alternative design choices. The best results are highlighted in \best{bold}.}
\label{tab:ablation_studies2}
\centering
\small
\begin{tabular}{lccc}
\toprule
\multirow{2}{*}{Alternatives} 
& \multicolumn{3}{c}{Metrics} \\
\cmidrule(lr){2-4}
& Dice$\uparrow$ & IoU$\uparrow$ & clDice$\uparrow$ \\
\midrule
Linear Fusion & 88.49\sd{7.83} & 80.17\sd{11.78} & 89.22\sd{8.19} \\
w/o $\mathcal{L}_{\mathrm{con}}$ & 88.38\sd{7.89} & 80.01\sd{11.81} & 89.03\sd{8.30} \\
w/o $\mathcal{L}_{1}$ & 85.93\sd{9.55} & 76.43\sd{13.37} & 87.23\sd{9.79} \\
RPP-SAM & \best{88.69}\sd{7.66} & \best{80.47}\sd{11.55} & \best{89.28}\sd{8.14} \\
\bottomrule
\end{tabular}
\end{table}

\begin{figure}[!t]
\centering
\includegraphics[width=0.97\columnwidth]{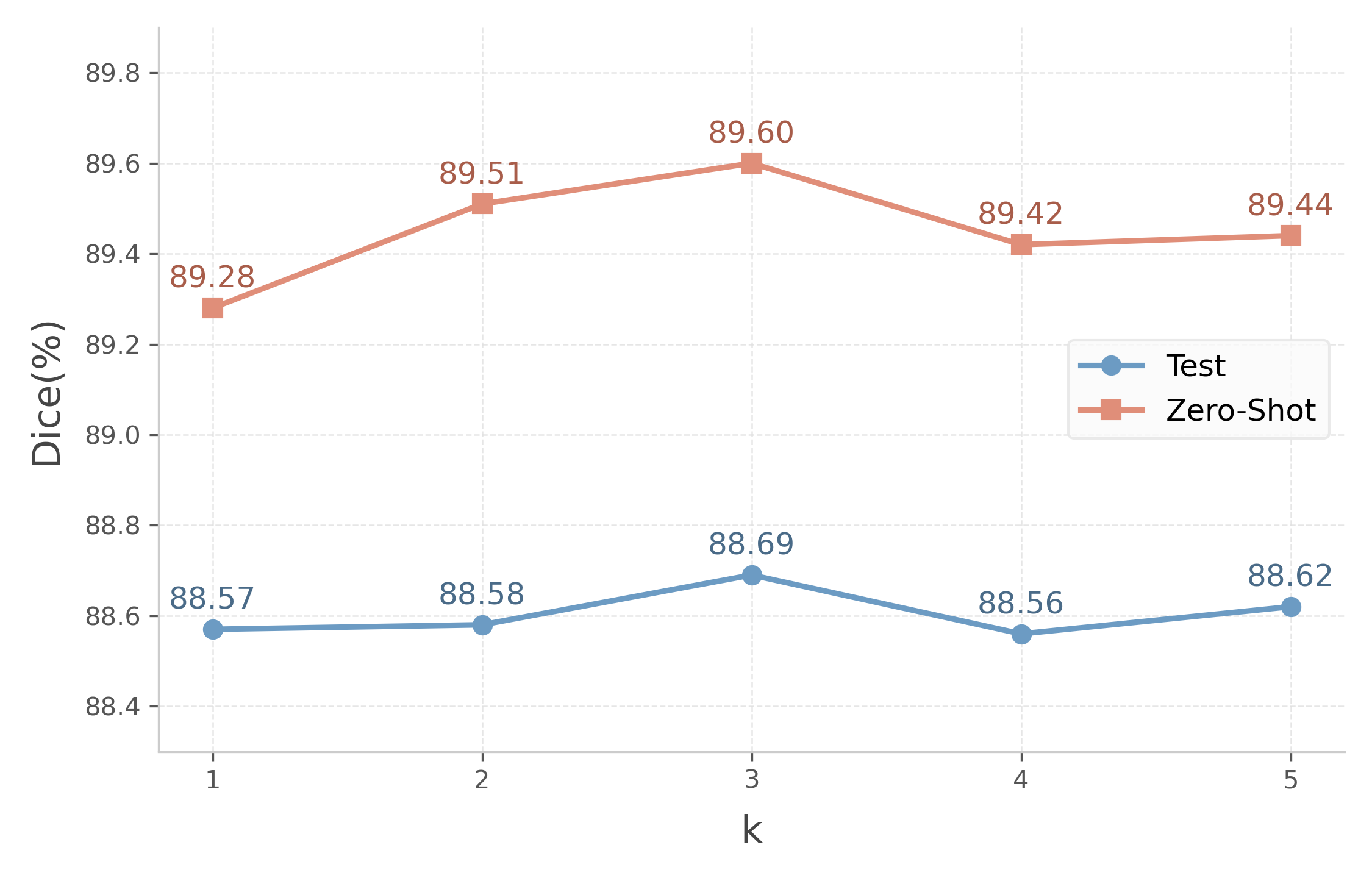}
\caption{Performance on the test and zero-shot test sets with different numbers of retrieved graphs ($k=1$--$5$).}
\label{fig:retrieval_k}
\end{figure}

\textbf{Choice of retrieval number $k$}.We investigated the sensitivity of RPP-SAM to the number of retrieved graphs, with $k$ ranging from 1 to 5. As shown in~\cref{fig:retrieval_k}, $k=3$ achieves the best performance on both the test and zero-shot test sets. Smaller values provide insufficient information for effective prototype learning, whereas larger values introduce potentially redundant or noisy structural cues, making prototype alignment more challenging and degrading segmentation performance.

\section{Conclusion}

In this paper, we propose RPP-SAM, a reference-guided progressive prototype-prompted framework for 2D vessel segmentation. To overcome the limitations of conventional prompting strategies and decoding architectures in SAM for vessel segmentation, we introduce vascular prompt embeddings (VPEs) and a dedicated vascular prototypical attention (VPA) mechanism to provide vascular-specific guidance throughout progressive decoding and recover fine-grained vascular structures. Furthermore, we propose reference-guided prototype learning to align mask-derived and reference-derived prototypes, enabling query-specific VPE generation without ground-truth masks at inference. Extensive experiments across 20 public datasets demonstrate that RPP-SAM consistently outperforms existing methods in overall segmentation performance, better recovers details such as thin vessels, and achieves strong zero-shot cross-dataset generalization.


\section{LLM Usage Statement}
Large language models (LLMs) and AI tools were used to assist with manuscript preparation and result visualization. Specifically, ChatGPT was used for language polishing and grammatical refinement, while Claude Code was used to assist with the refinement of result visualizations,~\cref{fig:dataset_performance} and~\cref{fig:segmentation}. All AI-assisted outputs were reviewed and verified by the authors, who take full responsibility for the content of this paper and the reported results.

{
    \small
    \bibliographystyle{ieeenat_fullname}
    \bibliography{main}
}


\end{document}